\documentclass[10pt, conference]{IEEEtran}

\IEEEoverridecommandlockouts                              

\usepackage{lipsum}
\usepackage{graphicx}                       
\usepackage{stfloats}
\usepackage[tight,footnotesize]{subfigure}  

\usepackage{amssymb,amsmath}
\usepackage{textcomp}
\usepackage{mdwmath}
\usepackage{mdwtab}
\usepackage{eqparbox}
\usepackage{rotating}                       
\usepackage{array}                          
\usepackage{listings}                       
\usepackage[ruled,vlined,noresetcount]{algorithm2e}
\usepackage{listings}
\usepackage[noend]{algpseudocode}

\title{\LARGE \bf Learning-Based Defect Recognitions for Autonomous UAV Inspections}

\author{Kangcheng Liu$^*$, Xiaodong Han, and Ben M. Chen
\thanks{$^{1}$K. Liu and B. M. Chen are with the Department of Mechanical and Automation Engineering, The Chinese University of Hong Kong, Shatin, N.T., Hong Kong. Kangcheng Liu is the corresponding author.
(email: {\tt\small kcliu@mae.cuhk.edu.hk, bmchen@cuhk.edu.hk)}} 
\thanks{$^{2}$X. Han is with the College of Computer and Control Engineering, Minjiang University, Fuzhou, China.
(email: {\tt\small hxdgod@mju.edu.cn)}}
}

\usepackage{hyperref}
\usepackage{footnote}
\usepackage{blindtext}
\usepackage{cleveref}

\begin{document}

\maketitle
\thispagestyle{empty}
\pagestyle{empty}

\begin{abstract}
Automatic crack detection and segmentation play a significant role in the whole system of unmanned aerial vehicle inspections. In this paper, we have implemented a deep learning framework for crack detection based on classical network architectures including Alexnet, VGG, and Resnet. Moreover, inspired by the feature pyramid network architecture, a hierarchical convolutional neural network (CNN) deep learning framework which is efficient in crack segmentation is also proposed, and its performance of it is compared with other state-of-the-art network architecture. We have summarized the existing crack detection and segmentation datasets and established the largest existing benchmark dataset on the internet for crack detection and segmentation, which is open-sourced for the research community. Our feature pyramid crack segmentation network is tested on the benchmark dataset and gives satisfactory segmentation results. A framework for automatic unmanned aerial vehicle inspections is also proposed and will be established for the crack inspection tasks of various concrete structures. All our self-established dataset and codes are open-sourced at: \href{https://github.com/KangchengLiu/Crack-Detection-and-Segmentation-Dataset-for-UAV-Inspection}{https://github.com/KangchengLiu/Crack-Detection-and-Segmentation-Dataset-for-UAV-Inspection}.

\end{abstract}

\section{Introduction}
Concrete structures play an important role in ground transportation networks. However, traditionally, concrete structure crack inspections are conducted manually by human operators with heavy and expensive mechanical equipment. It is logistically challenging, labor-intensive, costly, and dangerous, especially when inspecting the substructure and superstructure in harsh environments that are hard and dangerous to be accessed by human operators. Therefore, it is very meaningful and significant for us to develop a fully autonomous intelligent unmanned aerial system for inspecting large-scale concrete structures and detecting the defects such as cracks.

This study proposes a framework for automatic crack detection and segmentation for UAV inspections. The contribution of this study is fourfold.
\begin{enumerate}
 \item An overall unmanned aerial system architecture for UAV inspections is proposed.
 \item  We have established and tested neural network frameworks based on classical networks including Alexnet, VGG, and Resnet for crack detection.
 \item We have proposed and validated a hierarchical convolutional neural network (CNN) framework called \textit{CrackNet} which is efficient in crack segmentation on our self-established dataset. Comparisons with the other state-of-the-art segmentation network structures are also done.
 \item The benchmark dataset is merged and open-sourced for the research community.\footnote{\href{https://github.com/KangchengLiu/Crack-Detection-and-Segmentation-Dataset-for-UAV-Inspection}{https://github.com/KangchengLiu/Crack-Detection-and-Segmentation-Dataset-for-UAV-Inspection}} To our knowledge, this is the biggest dataset on the internet which can be used for both crack detection and segmentation so far. 
\end{enumerate}
The rest of the paper is organized as follows: Section II introduces the overall system architecture of our unmanned aerial systems. The deep learning architectures for crack detection are described in Section III. Section IV elaborately illustrates the feature pyramid network architectures for crack segmentation. Section V discusses the benchmark dataset setup and the crack segmentation experimental results. Section VI concludes the paper and proposes future works.   
\section{The System Architecture}

\subsection{The Avionic System}

\begin{figure}[htbp!]
    \centering
    \includegraphics[scale=0.25]{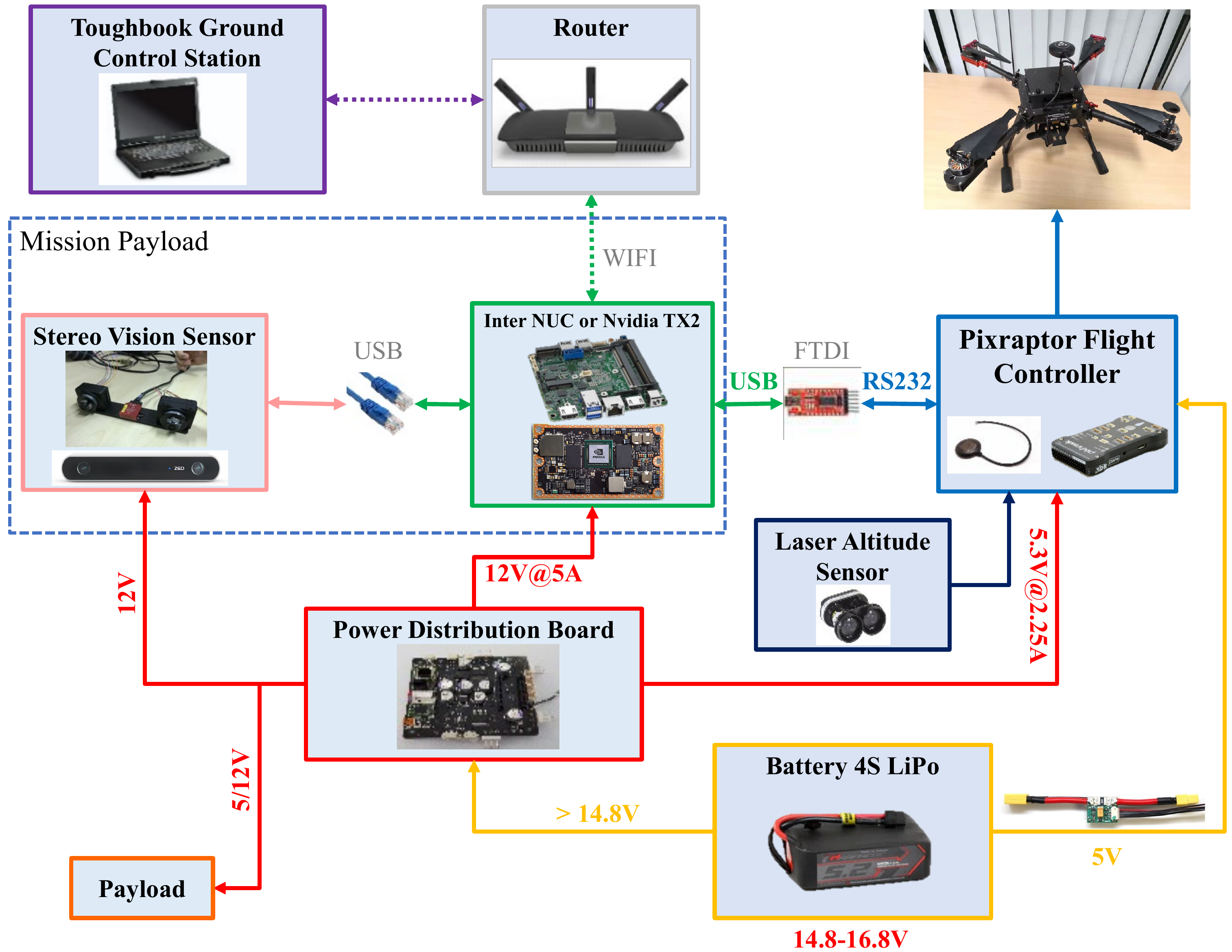}
    \caption{The avionic system}
    \label{fig:my_label}
\end{figure}
The overall hardware architecture of our quadrotor system is shown in Figure 1. The 450-sized UAV frame has been used for our UAV platform. A typical control method in our group is a cascade control loop tackling with the inner loop control (the attitude loop) and the outer loop control (the position loop), respectively. Among those control approaches the composite nonlinear feedback
(CNF) method for attitude stabilization and robust perfect tracking (RPT) method for trajectory tracking shows the best performance. We have used PX4 based flight controller to implement the flight control law. And we utilized Inter NUC (or Nvidia Jetson TX2 as an alternative) as the onboard mission computer for vision processing, target detection, etc. Communications of UAVs with other UAVs and GCS (Ground Control Station) are done using WIFI. 

\subsection{The Overall System for UAV Inspections}
\begin{figure}[htbp!]
\centering
\includegraphics[scale=0.265]{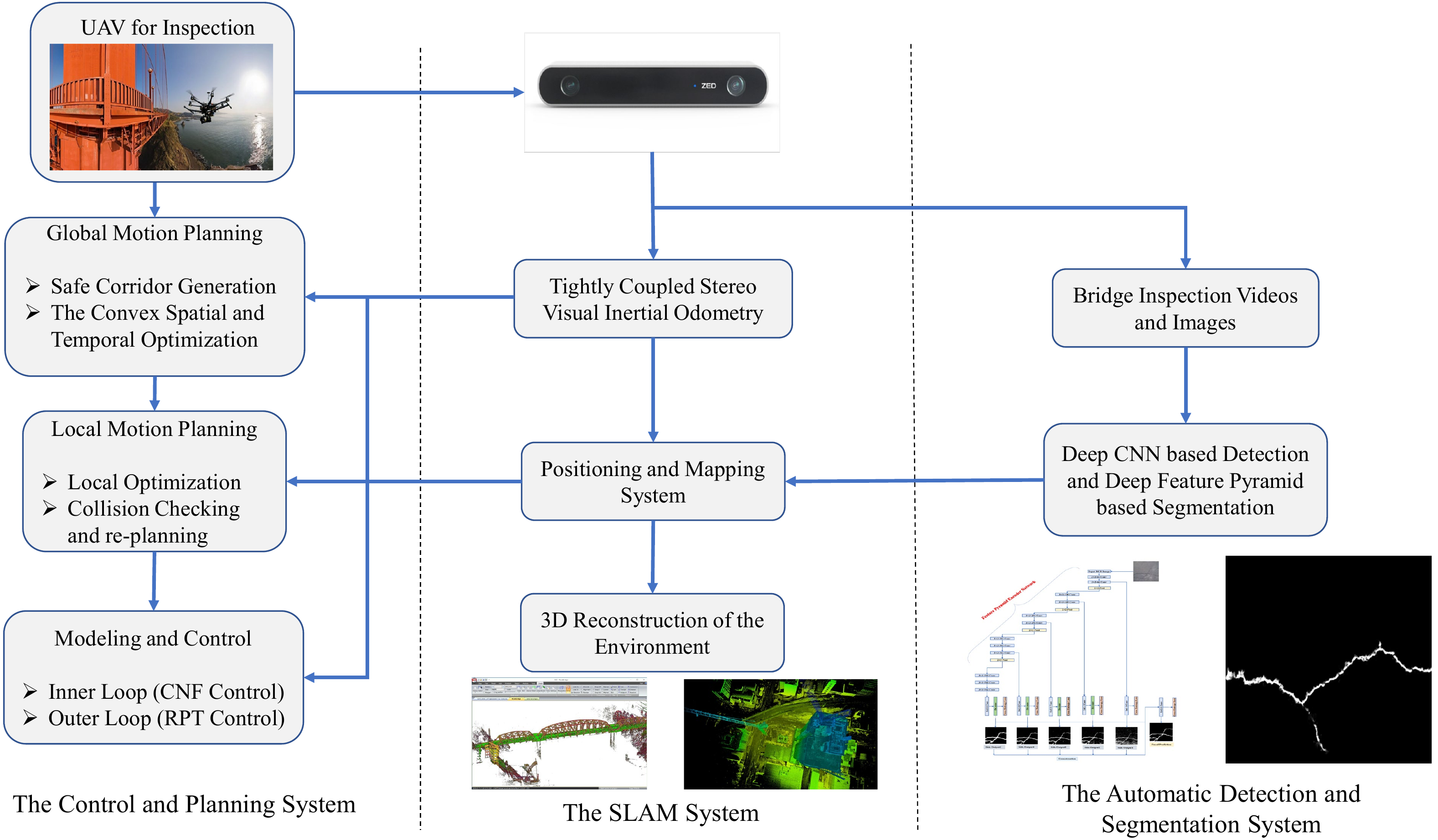}
\caption{The overall system structure}
\label{fig:my_label}
\end{figure}
As shown in Figure 2, the UAV inspection system consists of 3 subsystems: 
\begin{enumerate}
  \item The Control and Planning System 
  \item The Simultaneous Localization and Mapping (SLAM) System
  \item The Automatic Detection and Segmentation System
\end{enumerate}

The subsystems should be integrated to make an autonomous unmanned system, and the system integration of all the technical components is the most challenging task. Most importantly, a significant module for Unmanned Aerial Vehicle (UAV) intelligent inspection systems is to develop computer vision algorithms for processing images captured and detecting cracks and structural damages. The visual sensors have been emerging as novel techniques for modeling the environment \cite{liu2017avoiding, liu2020fg, liu2022weak, liu2021fg, liu2019deep, liu2022fg, zhao2021legacy, liu2022ARM, liu2022CYBER2, liu2022ICCA1, liu2022ICCA2,liu2022CYBER1, liu2022integrated, teamopenpcdet, team2020openpcdet}, and has been widely adopted for robotic applications. Various visual sensors and Light Detection And Ranging Sensors (LiDAR) sensors have been widely deployed in various robotics platforms such as automatically navigated aerial robots and ground robots \cite{liu2022industrial, liu2022d, yang2022datasets, yang2022datasetscbm, cao2021visdrone, liu2023lidar}. For example, the SLAM is widely adopted for robot localization, mapping, and navigation applications \cite{liu2022ICCA2, liu2022ws3d, yuzhi2020legacy, liu2022CYBER1, liu2022semicyber, liu2022lightarm, liu2022rm3d, liu2022weaklyeccv, liu2022lightarxiv, liu2022light, liu2022robustmm, liu2022integratedarxiv, liu2022ws3d}.

\section{Deep Learning Based Crack Detection}
In the past few years, CNN architecture has shown great performance in  detection and classification. A CNN architecture for classification typically consists of some convolutional blocks and several fully connected layers. Each convolutional block is composed of a convolutional layer, an activation unit, and a pooling layer. Considering the CNN architecture as a classifier, the crack detection problem can be formulated as a binary classification problem to classify between the crack and non-crack image samples.

\subsection{Dataset Preparation}
A representative and typical dataset is very important for the crack detection task. In the previous research, every method is validated on their self-established small-scale dataset to evaluate its performance, which makes the comparisons between different methods difficult. We have summarized different crack datasets and constructed a benchmark dataset for crack detection and segmentation. To our knowledge, this is the biggest dataset that can be utilized for both crack detection and segmentation and it will be beneficial for further research in this field.
\subsection{Data Preprocessing and Labeling}
The pre-processing of training images is done in order to make full use of the information in the image. The original big images are cropped into the 100$\times$100 sub-image samples and the sub-images are labeled as the crack or non-crack. We have used Photoshop and \textit{labelme} to obtain the pixel-level label of the image. If the percentage of the crack pixel in the cropped image is greater than 9\%, we define the image as the crack image.    

\subsection{The Network Structure and Training}
\begin{figure}[htbp!]
\centering
\includegraphics[scale=0.15]{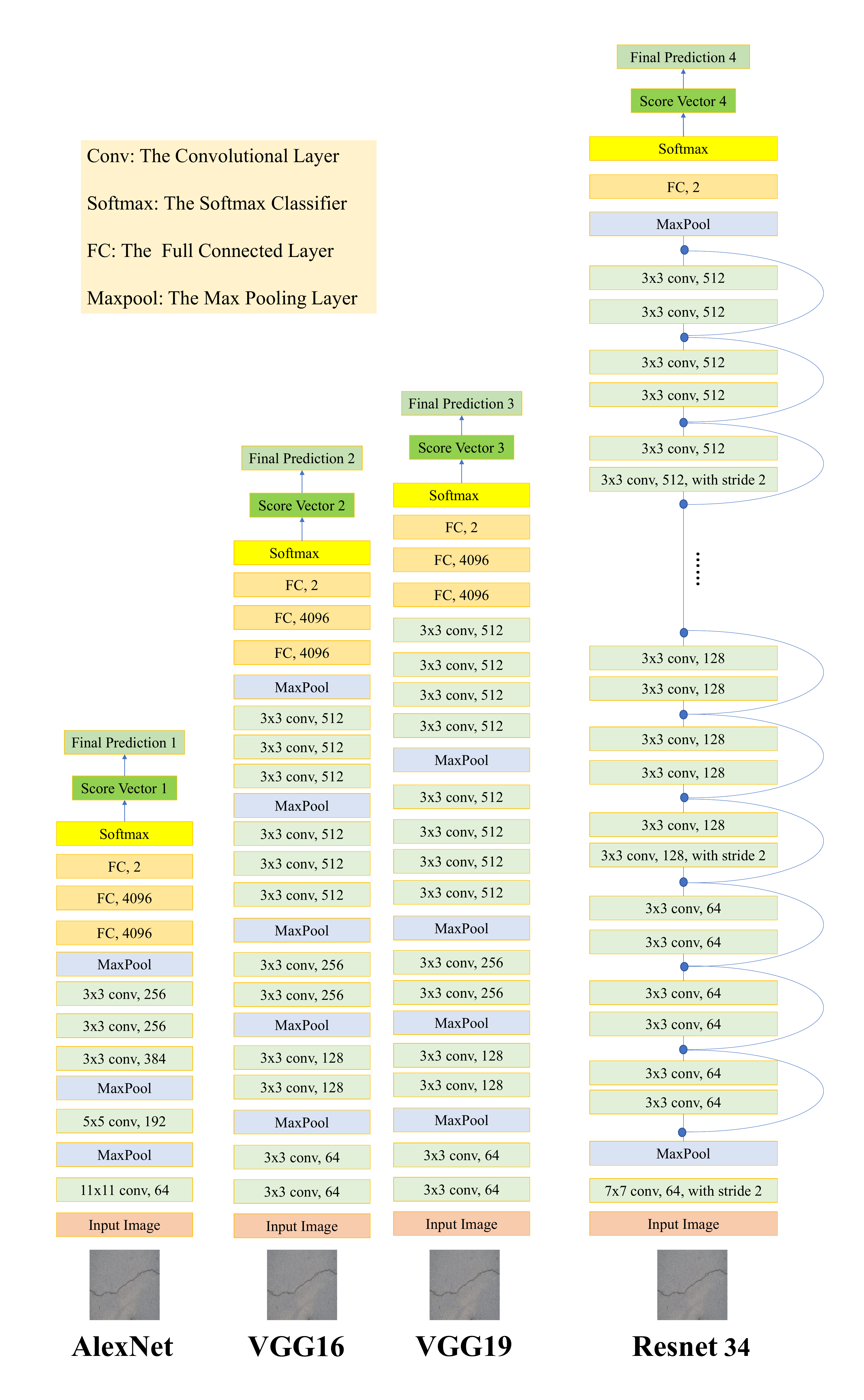}
\caption{The different network architectures to implement crack detection}
\label{fig:my_label}
\end{figure}
This paper used different architectures (VGG-net, Alexnet, Resnet) to do image classification and compare the results. The network architectures are shown in Figure 3. In order to accelerate the training process and improve the accuracy on our relatively small dataset (compared with \textit{ImageNet}), we utilize the transfer learning approach and use the pre-trained weight provided by \textit{Pytorch} to initialize the weights of different network architectures. 

The training runs 500 iterations with a batch size 32 over 15,000 images. The data is divided into 3 parts, 70\% for training, 15\% for testing and 15\% for validation. We set the learning rate as 1e-3 for the first 250 iterations and 3e-4 for the next 250 iterations. The cross-entropy loss was used for the network training. And the loss function can be formulated as follow:

\begin{footnotesize}
\begin{equation}
\begin{aligned}
L=\sum_{x^{In}} y(x^{In}) \cdot log(p(x^{In}))+ (1-y(x^{In})) \cdot log(1-p(x^{In}))]
\end{aligned}
\end{equation}
\end{footnotesize}
where $p(x^{In})$ represents the predicted possibility of a input image to be a crack image.  While $y$ is the label of the input image. For crack image, $y$=1. For non-crack image, $y$=0.

\section{Feature Pyramid Network Based Crack Segmentation}
The conventional CNNs architectures are more effective for image classification or recognition tasks, but not suitable for more complex image understanding tasks, such as semantic segmentation. The crack segmentation problem can be formulated as a pixel-level classification problem and the '0' and '1' refer to "non-crack" and "crack" respectively. Inspired by the hierarchical structure of Full Convolutional Network (FCN) and Segnet, we have proposed a feature pyramid-based CNN network structure for pixel-wise semantic segmentation.
\subsection{Motivation}
The main idea lies in that the high-level and low-level feature maps can be fused to obtain the final prediction feature map. Note that when doing upsampling on a convolutional feature map, the abstraction levels of the different convolution layers are different. The features learned by the shallow layer are local features. As the convolutional network becomes deeper, the receptive field continues to increase, the feature learned becomes more abstract. 

As shown in Table II and Figure 4, the optimal prediction occurs in the 2nd and 3rd convolutional feature map. It demonstrates that the shallow feature maps give a satisfactory prediction of crack profile but are very sensitive to local noise in image, and are often not robust when directly used for prediction. While the abstract features learned from deeper network layers are very robust to noise in image, but they are prone to suffer from gradient disappearance or boundary blurring, which makes the feature map too fuzzy to retaining the prediction boundaries for the crack. Therefore, in order to get accurate boundary segmentation, the tradeoff method is to upsample the feature maps of different levels to the same size and then fuse the outputs of different layers of the ConvNet to get a better final prediction feature map. 
\subsection{Network Architecture}
\begin{figure}[htbp!]
\centering
\includegraphics[scale=0.24]{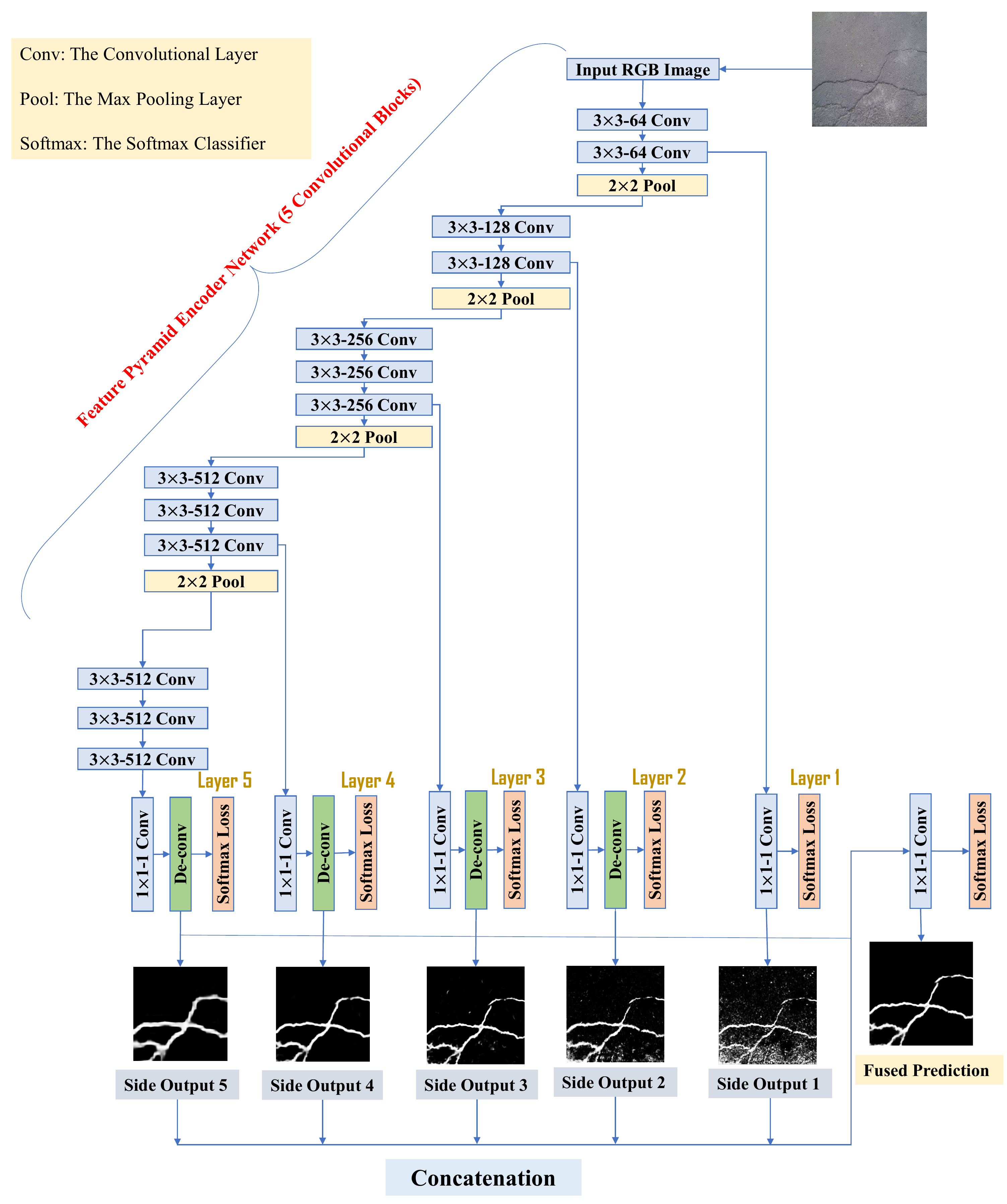}
\caption{The overall segmentation network structure}
\label{fig:my_label}
\end{figure}
As shown in the Figure 4, the first 13 convolutional layers are inspired by the VGG-16 network and they can be divided into 5 convolutional blocks. Batch normalization is used to speed up the convergence of the network training. Every convolutional layer is consisting of convolution, batch normalization and Rectified Linear Unit (ReLU). The fully connected layers are abandoned because pixel level prediction is required. And the last pooling layer is also not used because the output of it is too blurry to generate a satisfactory prediction feature map. The max-pooling of 2$\times$2 filter is adopted after each convolutional blocks of the VGG-16 backbone. It makes the amount of parameters in the network smaller while not causing the translational variance over the small spatial shift. Then we do 1$\times$1 convolution with one kernel to the output feature maps of the 5 convolutional blocks to make them become 1-channel. Next the \textit{bilinear Interpolation} is utilized to do deconvolution to the prediction feature map to make it become the same size as the input image. Finally, the linear fusion method can be used to concatenate outputs of different convolutional blocks. And 1$\times$1 convolution with 1 output depth is applied to the 5-channel concatenation result to make it become 1-channel fused prediction feature map. At last, the softmax classifier is used to do logistic regression for both side output prediction feature map and fused prediction feature map to get classification at pixel level. And the outputs of softmax classifier are 1-channel possibility maps indicating the possibility of each pixel to be a crack pixel. Then a fixed threshold can be assigned to get the prediction label based on the outputs of the softmax layer.

Finally, guided filtering (GF) post-processing is also utilized to do refinement to the prediction results.
\subsection{Loss Function Formulation and Optimization}
One of the major problems in crack segmentation is that more than 90$\%$ of the pixels are non-crack, which will easily cause training problems because of the imbalanced model updating. The original cross-entropy loss function can not be adopted because even if all the pixels are classified as non-crack, the prediction accuracy is still acceptable and correct prediction is given to most of the pixels. However, it makes no sense to our crack segmentation task. Therefore, rather than simply use the cross-entropy loss function, we adopted the class-balanced loss function to weigh between the crack and non-crack pixel differently, which is inspired by previous works. The details will be discussed as follows:

Denote the training set which includes $N$ images as $V={(S^n, T^n), n=1, 2,..., N}$, in which the set $S^n=\{{s_j^{(n)}, j=1, 2,..., J}\}$ represents the original input image, the set $T^n=\{{t_j^{(n)}, j=1, 2,..., J, t_j^{(n)} \in \{0,1\}}$\} represents the ground truth crack label feature map which is corresponding to $S^n$. $J$ denotes the total amount of pixels in each image. For each labeled pixel, the target is to train our network to learn a model that produces a final prediction which is consistent with the ground truth. Denote $H$ as the number of convolutional blocks ($H= 5$ in our case), the feature map generated by the side output of each convolutional block can be written as $F^{(h)}={F_j^{(h)}, j= 1,.., J}$, where $h=1, 2,.., H$. And the fused feature map can be written as $F^{fused}$=$F_j^{fused}, j=1, 2,.., J$. Then the parameters of the convolutional layers of VGG net can be defined as $W_1$, and the parameters of the sided output convolutional layers and final fuse convolutional layer can be defined as $W_2=\{w_{1},w_{2},...,w_{H}, w_{fuse}\}$, in which $H$ is the number of the side output, which is equal to the number of convolutional blocks in our case. The total parameter set of ${W_1, W_2}$ can be denoted as ${W}$. For the crack segmentation task, we should attach less significance to the loss function of non-crack pixels which is in the majority, while attach more significance to the loss function crack pixels which are in the minority. We adopted median frequency balancing in the weight determination. If the total number of crack and non-crack pixels in the training set are $p$ and $q$ respectively. The class frequencies of crack and non-crack are $\frac{p}{p+q}$ and $\frac{q}{p+q}$, while the median for the 2 classes is 0.5. Then the median divided by the class frequency gives the weight of 2 classes. In our case, the weights of the loss function for the crack pixels and non-crack pixels are $\alpha_1=\frac{p+q}{2p}$ and $\alpha_2=\frac{p+q}{2q}$ respectively. Then for each side-output layer, the improved loss function can be formulated as:

\begin{footnotesize}
\begin{equation}
\begin{aligned}
L_{side}^{h}(F_j; W_1; W_2)=-&\alpha_2\sum_{j\in  S^{-}} log(1-P(F_j; W_1; W_2))\\-&\alpha_1\sum_{j \in S^{+}} log(P(F_j; W_1; W_2))
\end{aligned}
\end{equation}
\end{footnotesize}
where $h=1, 2..., H$ respectively, $S^{+}$ and $S^{-}$ are the total number of crack pixels and non-crack pixels respectively for an input image. And \textit{P} denotes the predicted possibility of each pixel to be a crack one. Next the improved loss function for the fused output can be also written as:

\begin{footnotesize}
\begin{equation}
\begin{aligned}
L_{fuse}(F_{j}^{fused}; W_1; W_2)=-&\alpha_2\sum_{j\in  S_{-}}log(1-P(F_j; W_1; W_2))\\-&\alpha_1\sum_{j\in  S_{+}}log(P(F_j; W_1; W_2))
\end{aligned}
\end{equation}
\end{footnotesize}

And then the total loss function is written as:

\begin{footnotesize}
\begin{equation}
\begin{aligned}
L_{total}(W_1; W_2)=\sum_{j=1}^{J}(\sum_{h=1}^{H}\lambda_hL_{side}^{h}(F_j; W_1; W_2)\\+L_{fuse}(F_{j}^{fused}; W_1; W_2))
\end{aligned}
\end{equation}
\end{footnotesize}

The parameter $\lambda_h$ can be tuned to assign different weights to the loss at different levels of the feature map. And the weights are finally set as $\{\lambda_1, \lambda_2, \lambda_3, \lambda_4, \lambda_5\}$ = $\{0.5, 1.0, 0.8, 0.5, 0.3\}$, which gives the best segmentation results. The detailed experiment results and comparisons are shown in Section V.
Finally, the computed total loss function can be utilzed for back propagation in the optimization procedure which is similar to Stochastic Gradient Descent (SGD). To improve the convergence speed, Adam algorithm was adopted to achieve SGD with momentum which accumulates the past gradient to accelerate the gradient downhill.

\subsection{Model Parameters and Data Argumentation}
The network proposed above is implemented on the open-source deep learning framework \textit{Pytorch}. The parameters of the whole convolutional network are initialized with normal distribution. And the \textit{Bilinear Interpolation} is utilized to do upsampling in the deconvolution process. The learning rate is set to 9e-5 and the batch size is set to 2. The momentum and weight decay are set to 0.8 and 6e-4 respectively in the gradient descent optimization. The network is trained with 2.4e5 iterations in total and the learning rate will be decreased by 20\% after every 3e4 iterations. All the training are done on an Nvidia GeForce GTX RTX2080Ti GPU.

Data argumentation also plays a significant role in the training of the networks to enhance the utilization of the training dataset. The images are rotated to 6 different angles from 0-degree to 360-degree with a 60-degree interval. And we also flipped the images horizontally therefore the whole dataset is enhanced 12 times. In this way, we can significantly increase the diversity of data available for training models, without actually collecting new data. 

\section{Experimental Results}
\subsection{Comparison of Results of crack detection}
\begin{table}[htbp!]
\caption{The comparison of detection results between methods}
\label{table_example}
\begin{center}
\begin{tabular}{c|c|c}
\hline
Network Architecture & Accuracy/\% & Validation Time /ms\\ 
\hline
Alexnet &94.5& 700.2\\ 

VGG 16 &95.6& 678.5\\

VGG 19  &95.4& 689.6\\

Resnet  &96.8& 617.5\\
\hline
\end{tabular}
\end{center}
\end{table}
Crack detection test was done on the device with Intel Xeon E5-2620 v4 CPU and Nvidia RTX 2080Ti GPU. As shown in Table I, the Resnet has the best accuracy and interference speed among different kinds of networks. Although the CNN architectures can not give perfect prediction, the result for crack detection is important because it can be utilized for coarse localization of the defected region. And the definition for crack sub-images based on the percentage of the crack pixel is not always accurate. More accurate prediction will require human to label the crack or non-crack sub-image ground truth which is time-consuming and labour-consuming. Therefore segmentation of crack at pixel level is necessary for a more complete crack inspection system. 

\subsection{The Benchmark Dataset Setup}
In our project, efforts have been made to merge more than 10 different crack segmentation datasets available on the Internet. And we also collected some crack images from taking photos of concrete structures and labeled them utilizing an effective graphical image annotation tool: \textit{labelme}. The improvement can be made based on the benchmark dataset and there is no doubt that this will be beneficial and crucial to the whole research community of crack detection and segmentation. In addition, more data will be added and merged into the dataset with the research project going on. 
Following are some of the datasets which we collected images from:
\subsubsection{GAPs Dataset}
The German Asphalt Pavement Distress (\textit{GAPs}) dataset provides a standardized large-scale and high-quality dataset. A total of 1969 gray-scale images are included in the GAPs dataset.

\subsubsection{CRACK500 Dataset}
The \textit{CRACK500} is a dataset containing more than 500  $2,000 \times 1,500$ images by using phones to collect data. A pixel-level labeled binary map is assigned to each image.

\subsubsection{CFD Dataset}
 The \textit{CFD} is a crack dataset containing 118 480 $\times$ 320 images. The images are captured by iPhone and the profiles of the cracks are manually annotated.

\subsubsection{AELLT Dataset}
The \textit{AELLT} is a crack patterns dataset containing 58 crack images of 256 $\times$ 256 pixels. The contours of the crack are annotated therefore it can be utilized for crack segmentation. 

\subsubsection{Cracktree260 Dataset}
The \textit{cracktree260} is a dataset containing 260 images of size 800 $\times$ 600 with various cracks. The annotation of this dataset gives pixel-wise label, which is very convenient to be used directly for training and testing.

After adding our own data, the crack segmentation dataset consists of approximately 11300 images. To the best of our knowledge, this is the biggest dataset for crack segmentation on the internet so far. All the images are resized to 256 $\times$ 256 for training purpose. The final training set consists of 9000 images and the testing set consists of 1500 images.

\subsection{The Evaluation Metrics}
In order to evaluate the performance of the segmentation, we adopted typical metrics of the semantic segmentation community to evaluate our results. The metrics are listed as follows:
\begin{itemize}
    \item The Global Pixel Accuracy (A), which is defined as the percentage of the correctly predicted pixel. 
    \item The Mean Intersection Over Union (MIOU), which is a significant metric in semantic segmentation.
    \item The best $F$-measure on a dataset based on a fixed threshold (DS)
    \item The aggregated $F$-measure score on a dataset for the best threshold in images (IS)
    \item The precision at the best $F$-measure DS (BP)
    \item The recall at the best $F$-measure DS (BR)
 \end{itemize}
The detailed definition of them are shown as follows: 
The global pixel accuracy (A) and mean intersection over union  (MIOU) are well-defined in.
Denote the number of true positive, true negative, false positive, false negative samples as $N_{TP}$, $ N_{TN}$, $ N_{FP}$ and $ N_{FN}$ respectively, the precision (P) can be represented as: 

\begin{footnotesize}
\begin{equation*}
   {\rm P}=\frac{N_{TP}}{N_{TP}+N_{FP}}
\end{equation*}
\end{footnotesize}
And the recall (R) can be represented as:

\begin{footnotesize}
\begin{equation*}
  {\rm R}=\frac{N_{TP}}{N_{TP}+N_{FN}}
\end{equation*}
\end{footnotesize}
Then the $F$-measure can be represented as: 

\begin{footnotesize}
\begin{equation*}
 F-{\rm measure} = \frac{2PR}{P+R}
\end{equation*}
\end{footnotesize}
And the best $F$-measure on a dataset for a fixed threshold \textit{m} (DS) can be written as:

\begin{footnotesize}
\begin{equation*}
 {\rm
DS}=max\{\frac{2P^m \times R^m}{P^m+R^m}, :  m=0.01, 0.02, 0.03,..., 0.99\}
\end{equation*}
\end{footnotesize}

 The precision and recall at the best $F$-measure are denoted as BP (Best Precision) and BR (Best Recall) respectively. 
 While the aggregated $F$-measure score on a dataset for the best scale in the image (IS) can be given as:
 
 \begin{footnotesize}
 \begin{equation*}
  {\rm IS
 }= \frac{1}{N_{im}} \sum_{i=1}^{N_{im}}max\{\frac{2P^m_i\times R^m_i}{P^m_i+R^m_i}: m=0.01, 0.02, 0.03,..., 0.99\}
 \end{equation*}
 \end{footnotesize}
 
 where $m$ denotes the threshold, $i$ denotes the index of the image and $ N_{im}$ denotes the total amount of images. $ P_m$ and $R_m$ are the precision and recall at the threshold $ m$. $ P^m_i$ and $ R^m_i$ denote the precision and recall of image $i$.
\subsection{The Comparison of Segmentation Results Between Different Side Outputs}

\begin{table}[htbp!]
\caption{The comparison of segmentation results between different side outputs}

\label{table_example}
\begin{center}
\begin{tabular}{c|c|c|c|c|c|c}
\hline
Output& A & MIOU& BP & BR& DS& IS\\
\hline
Side 1&94.9&82.8&81.1&79.8&80.4&81.1\\ 

Side 2&95.6&85.3&85.3&81.7&83.5&84.2\\

Side 3&95.2&84.7&83.2&82.8&83.0&83.8\\

Side 4&94.7&82.2&78.3&81.0&79.6&80.9\\

Side 5&91.8&76.5&66.3&77.7&71.5&72.3\\

\textit{CrackNet}&96.3&86.8&84.8&84.6&84.7&85.8\\
\textit{CrackNet} (With GF)& \textbf{96.8}&\textbf{87.7}&\textbf{86.6}&\textbf{84.8}&\textbf{85.7}&\textbf{86.5}\\
\hline
\end{tabular}
\end{center}
\end{table}
Firstly, we did a visualization of different convolutional layers and compared the segmentation results. The final segmentation prediction of different side output layers, fused and GF refined results are shown in Figure 5. The side  outputs of the shallower layers give a fine prediction of the crack contour, but are sensitive to noises. While the outputs of the deeper layers can successfully filter the local noises in the image, but the contours for the cracks are blurred. As shown in Table II, the best segmentation result of the sides output feature maps is given by the side output of 2nd and 3rd layer. And the fused result with guided filter post-processing gives the best segmentation performance if trained with the best loss function weights at different level, which will be discussed below in Subsection E.

\begin{figure}[htbp!]
\centering
\includegraphics[scale=0.225]{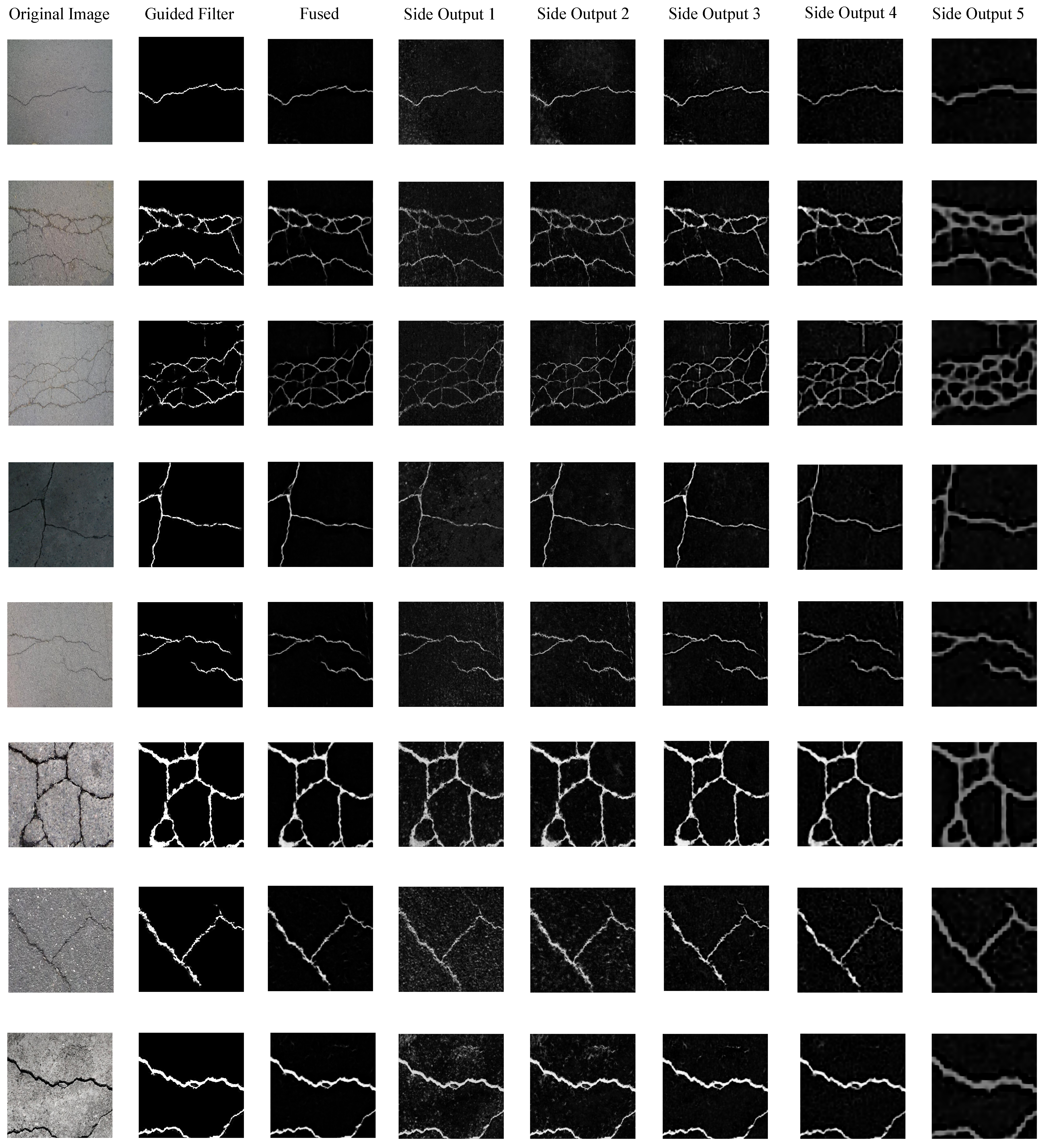}
\caption{Segmentation results of the side outputs, fused results and guided filter results}
\label{fig:my_label}
\end{figure}

\subsection{The Comparison of Loss Function Weights at Different Level}
The hyper-parameter setting in the loss function to weight losses between different side outputs is very important to give a better semantic segmentation  result. However, it is impossible to try all combinations of parameters so we choose some typical parameters settings to evaluate the influence of them on final prediction results. In the first 2 cases, we give larger weights to the shallower layers, while in the 3rd and 4th cases, larger weights are given to the deeper layers. In the 6th and 7th cases, largest weights are given to the 2nd layer and 3rd layer, which gives the best feature map prediction among different layers. While in the 5th case, the weights are equal among different layers. The parameter setting of different cases are:

\begin{itemize}
    \item Case 1: $\{\lambda_1, \lambda_2, \lambda_3, \lambda_4, \lambda_5\}=\{4.0, 2.0, 1.0, 0.5, 0.25\}$
    \item Case 2: $\{\lambda_1, \lambda_2, \lambda_3, \lambda_4, \lambda_5\}=\{9.0, 3.0, 1.0, \frac{1}{3}, \frac{1}{9}\}$
    \item Case 3: $\{\lambda_1, \lambda_2, \lambda_3, \lambda_4, \lambda_5\}=\{0.25, 0.5, 1.0, 2.0, 4.0\}$:
    \item Case 4: $\{\lambda_1, \lambda_2, \lambda_3, \lambda_4, \lambda_5\}=\{\frac{1}{9}, \frac{1}{3}, 1.0, 3.0, 9.0\}$:
    \item Case 5: $\{\lambda_1, \lambda_2, \lambda_3, \lambda_4, \lambda_5\}=\{1.0, 1.0, 1.0, 1.0, 1.0\}$:
    \item Case 6: $\{\lambda_1, \lambda_2, \lambda_3, \lambda_4, \lambda_5\}=\{0.3, 0.7, 1.0, 0.7, 0.3\}$:
    \item Case 7: $\{\lambda_1, \lambda_2, \lambda_3, \lambda_4, \lambda_5\}=\{0.5, 1.0, 0.8, 0.5, 0.3\}$:
\end{itemize}

\begin{table}[htbp!]
\caption{The comparison of assigning different weights to the loss at different level of feature map}
\label{table_example}
\begin{center}
\begin{tabular}{c|c|c|c|c|c|c}
\hline
Case& A & MIOU & BP & BR& DS& IS\\
\hline
Case 1&94.4&83.1&82.1&80.8&81.4&82.9\\ 

Case 2&94.5&82.8&81.9&81.2&81.5&82.4\\

Case 3&94.6&83.3&82.2&81.5&81.8&83.3\\

Case 4&94.7&83.8&82.7&81.2&81.9&82.9\\

Case 5&95.8&86.5&84.3&83.8&84.0&84.5\\

Case 6&96.2&86.9&85.6&84.3&84.9&85.7\\

Case 7 \textbf{\textit{(best)}}&\textbf{96.8}&\textbf{87.7}&\textbf{86.6}&\textbf{84.8}&\textbf{85.7}&\textbf{86.5}\\
\hline
\end{tabular}
\end{center}
\end{table}

As shown in Table III, from the results, we can see that giving larger weights to the deeper layers or shallower layers did not contribute to improve the segmentation performance. While giving larger weight to the 2nd and 3rd layer gives the best performance. This is corresponding to the fact that the best feature map prediction is given by the 2nd and 3rd layers. And more significance should be given to the loss function of them to give a better segmentation performance.

\subsection{The Comparison with other Methods}

In order to verify the effectiveness of our method, the results of our \textit{CrackNet} is compared with 4 other very popular state-of-art CNN architectures in semantic segmentation.
\subsubsection{HED}
The \textit{HED} is a multi-level feature learning network which shows great performance in edge detection. 
\subsubsection{Segnet}
The \textit{Segnet} is a deep convolutional encoder-decoder architecture for Image Segmentation.
\subsubsection{FCN}
The \textit{FCN} is a fully connected network which also shows its performance in semantic segmentation.
\subsubsection{U-Net}
The \textit{U-Net} is a network which also has the encoder-decoder architecture, and it is initially utilized for biomedical image segmentation. 
\begin{figure}[htbp!]
\centering
\includegraphics[scale=0.20]{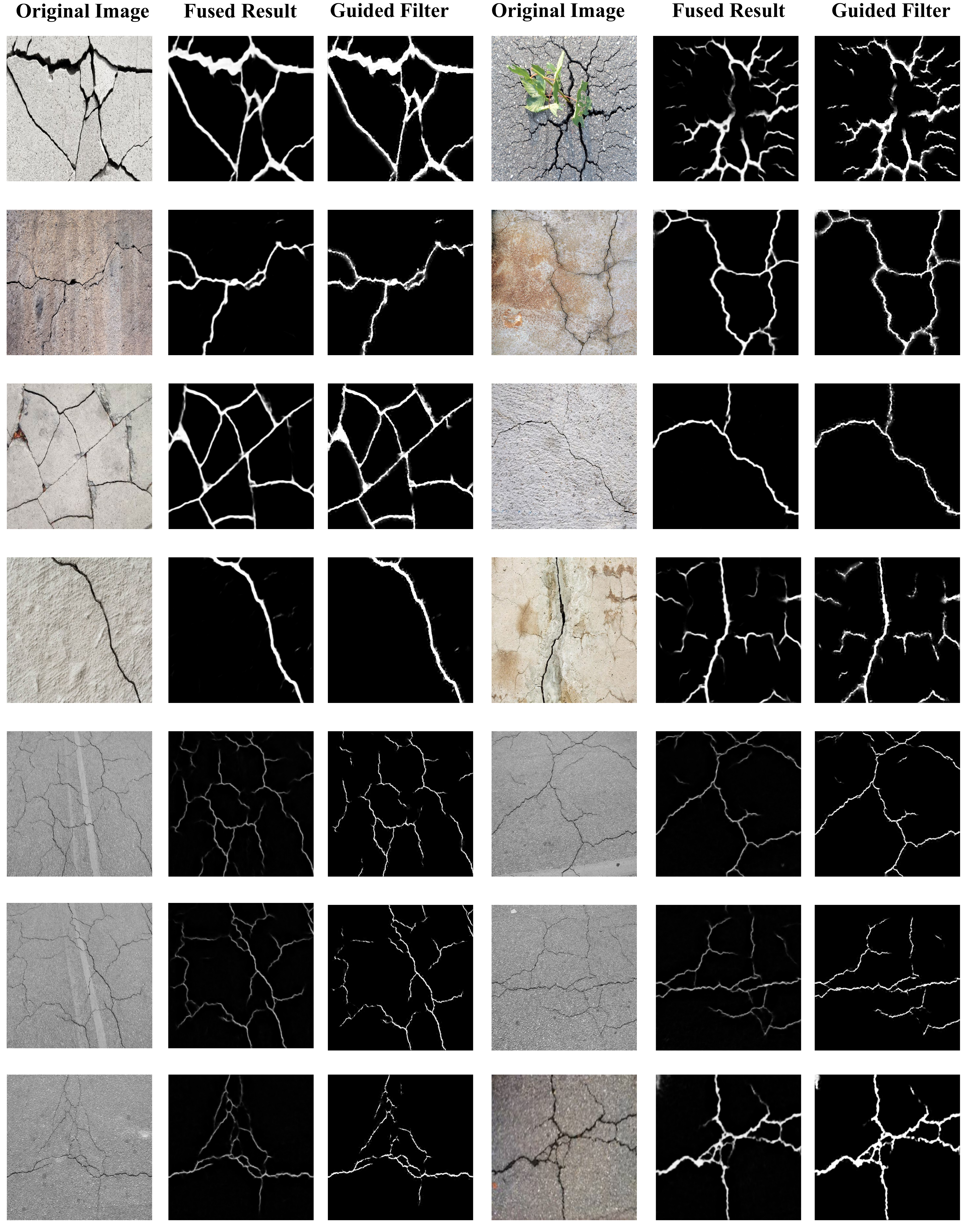}
\caption{Segmentation results in some challenging circumstances with various cracks and noises}
\label{fig:my_label}
\end{figure}

\begin{table*}[htbp!]
\caption{The comparison of segmentation results between methods}
\label{table_example}
\begin{center}
\begin{tabular}{c|c|c|c|c|c|c|c|c}
\hline
Methods& Inference Times/(ms)& Threshold (\textit{m}) & A & MIOU & BP & BR& DS& IS\\
\hline
HED & 67&0.47&92.8&75.9&74.5&79.3&76.8&77.5\\ 

Segnet &192&0.55&96.0&85.5&83.3&84.8&84.0&84.6\\

FCN-8s & 105&0.57&93.1&75.3&74.2&75.5&74.8&75.9\\

U-Net & 102&0.56&96.1&85.9&84.0&84.6&84.3&85.1\\

\textit{CrackNet} (Ours)& 132&0.48&\textbf{96.8}&\textbf{87.7}&\textbf{86.6}&\textbf{84.8}&\textbf{85.7}&\textbf{86.5}\\

\textit{Dialated-CrackNet} (Ours)& 141&0.49&\textbf{97.0}&\textbf{87.8}&\textbf{86.5}&\textbf{85.3}&\textbf{85.9}&\textbf{86.7}\\
\hline
\end{tabular}
\end{center}
\end{table*}


\begin{table*}[htbp!]
\caption{The comparison of segmentation results between methods}
\label{table_example}
\begin{center}
\scalebox{1.0}{\begin{tabular}{c|c|c|c|c|c|c|c|c}
\hline
Methods& Inference Times/ (ms)& Threshold (\textit{m}) & Average Precision & MIOU & Precision & Recall & DS& IS\\

\hline
HED & 97&0.47&92.8&75.9&74.5&79.3&76.8&77.5\\ 

Segnet &222&0.55&96.0&85.5&83.3&84.8&84.0&84.6\\

FCN-8s & 135&0.57&93.1&75.3&74.2&75.5&74.8&75.9\\

U-Net & 132&0.56&96.1&85.9&84.0&84.6&84.3&85.1\\

DeepLabV3 & 130&0.56&96.2&87.2&84.9&84.9&84.7&85.8\\

DeepLabV3+  & 132&0.55&96.3&87.8&85.4&85.5&85.1&85.9\\

PSPNetV1 & 130&0.57&96.2&87.9&85.5&85.6&85.3&86.3\\

SDD-Net & 142&0.57&96.2&87.9&85.5&85.6&85.3&86.3\\

\textit{CrackNet} (Ours)& 162&0.48&\textbf{96.8}&\textbf{87.7}&\textbf{86.6}&\textbf{84.8}&\textbf{85.7}&\textbf{86.5}\\

\textit{Dialated-CrackNet} (Ours)& 171&0.49&\textbf{97.0}&\textbf{87.8}&\textbf{86.5}&\textbf{85.3}&\textbf{85.9}&\textbf{86.7}\\
\hline
\end{tabular}}
\end{center}
\end{table*}

Our \textit{CrackNet} is specially designed for crack segmentation with pre-processing and post-processing methods as well as fine tuned weight given in Section IV. From the results in Table IV, it is demonstrated that \textit{CrackNet} with guided filtering post-processing and the best loss function weight gives better performance than other methods in crack segmentation task. As shown in Figure 6, the detection results in some challenging circumstances with various cracks and noises also demonstrate the effectiveness and robustness of our method.

\section{Conclusions and Future Work}
In this work, we have proposed a formal and systematic solution to automatic crack detection and segmentation for UAV inspections. The system architecture for UAV inspections is proposed and the network architecture for crack detection and segmentation is developed. More importantly, a benchmark dataset is also established and open-sourced for the community. In the future, we plan to concentrate on the integration of the sub-components and subsystems of the unmanned aerial system. More typical images of the concrete cracks will be merged into the dataset with our research project going on. The concrete defects are not restricted to cracks and specific inspection techniques should be developed for other structural damages. With the advancement of CNN architecture for semantic segmentation, we also plan to develop more advanced CNN architectures for better performance in the crack detection and segmentation task. As a finished \cite{liu2022enhanced} work, the developed can be adapted for autonomous UAV inspections for various modern infrastructures.

\addtolength{\textheight}{0cm}   





\bibliographystyle{IEEEtran}
\bibliography{references}

\end{document}